%% file: acl_latex.tex
\pdfoutput=1

\documentclass[11pt]{article}

\usepackage[final]{acl}

\usepackage{times}
\usepackage{latexsym}
\usepackage{amsmath, amssymb}
\usepackage{algorithm}
\usepackage{algpseudocode}
\usepackage{bm}
\usepackage{comment}

\usepackage[T1]{fontenc}

\usepackage[utf8]{inputenc}

\usepackage{comment}
\usepackage{microtype}

\usepackage{inconsolata}

\usepackage{graphicx}

\usepackage{xspace}
\usepackage{colortbl}

\newcommand{\ours}{Gold-Switch\xspace}

\usepackage{etoolbox} %

\newtoggle{showprev}
\toggletrue{showprev}

\newtoggle{shownew}
\toggletrue{shownew}

\usepackage{array} %
\makeatletter
\newcommand*{\@rowstyle}{}
\newcommand*{\rowstyle}[1]{%
  \gdef\@rowstyle{#1}%
  \@rowstyle\ignorespaces%
}
\newcolumntype{=}{%
  >{\gdef\@rowstyle{}}%
}
\newcolumntype{+}{%
  >{\@rowstyle}%
}
\makeatother

\definecolor{Emerald}{RGB}{192, 41, 66}
\iftoggle{shownew}{
    
    \newcommand{\jsleeobj}[1]{ %
    {\let\Cap\caption
    \def\caption##1{\Cap{\color{Emerald}##1}}
    \color{Emerald}#1}
    }
}
{
    
    \newcommand{\jsleeobj}[1]{#1}
    
}
\definecolor{Gray}{RGB}{200, 200, 200}
\iftoggle{showprev}{
    \newcommand{\jsleeprev}[1]{{\color{Gray}#1}}
    \newcommand{\jsleeprevobj}[1]{ %
    {\let\Cap\caption
    \def\caption##1{\Cap{\color{Gray}##1}}
    \color{Gray}#1}
    }
    \newcommand{\jsleeprevrow}[1]{\rowstyle{\color{Gray}}#1}
}
{

    \newcommand{\jsleeprev}[1]{}
    \newcommand{\jsleeprevobj}[1]{}
    \newcommand{\jsleeprevrow}[1]{}
}

\title{\ours: Training-Free Superposition of Slow- and Fast- Thinking LLMs}

 \author{\textbf{
Jaeseong Lee\thanks{~~Equal contribution},
Dayoung Kwon\footnotemark[1],
Seung-won Hwang\thanks{~~Correspond to seungwonh@snu.ac.kr}}\\
  {Computer Science and Engineering, Seoul National University} \\
  \texttt{\{tbvj5914,ekdud1660,seungwonh\}@snu.ac.kr} 
 }

\begin{document}
\maketitle
\begin{abstract}

\input{contents/0_abs}
\end{abstract}

\input{contents/1_intro}

\input{contents/2_rel}

\input{contents/3_prop}

\input{contents/4_exp}

\input{contents/5_concl}

\bibliography{2025NIPS_autogen.bib,custom.bib,ThinkSwitch.bib}

\clearpage
\appendix
\input{contents/7_appendix}

\end{document}

%% file: contents/0_abs.tex
 Large Reasoning Models (LRMs) excel in structured tasks by emulating deliberate human reasoning but often suffer from overthinking, degrading performance and wasting resources. One possible baseline is to deploy both LLM and LRM, then route input by predicting whether it requires reasoning and may cause overthinking.
 However, deploying multiple models
 can be costly or impractical. We propose a superposed deployment strategy with a lightweight, training-free regulation to optimize inference by switching one model on and off. Instead of routing, we selectively unlearn from LRM at inference, scaling down computation while preserving reasoning. By analyzing the cumulative energy of singular values, we identify optimal low-rank projections to adjust reasoning just right.

%% file: contents/1_intro.tex
\section{Introduction}
Recently, large reasoning models (LRMs), such as DeepSeek-R1~\citep{DeepSeekR12025deepseek-ai}, or QwQ~\citep{qwq32b}, have emerged as a powerful class of reasoning-oriented models. These models, starting from a base large language model (LLM), then trained with extensive chain-of-thought (CoT) supervision, demonstrate strong performance across complex reasoning tasks such as mathematical problem solving, programming, and multi-step question answering. 
LRMs exemplify a new generation of LLMs optimized for ``slow thinking''-- producing elaborate intermediate steps that mimic human reasoning. 

However, overthinking~\citep{Stop2025sui,NOT2025chen,Danger2025cuadron} has recently been identified as a critical challenge in LRMs: LRMs often generate unnecessarily long or convoluted reasoning paths~\citep{Stop2025sui}. 
This behavior has been shown to reduce both efficiency and even lower the task performance~\citep{Danger2025cuadron}. 
This highlights the need to switch between slow- and fast- thinking in LRMs.

One possible solution is training a model to allow switching between slow- and fast- thinking by predefined prompt, such as Qwen3~\cite{Qwen32025yang}.
However due to their limited performance, the recent trend is back to releasing LRM and LLM separately-- models such as Qwen3-2507, Qwen3-VL, Qwen3-Next~\cite{Qwen32025yang}, Qwen3-Omni~\cite{Qwen3Omni2025xu}, and Qwen3-Max~\cite{qwen3max} all trained and released slow- and fast- variants separately.

An alternative solution 
is deploying both LRM and LLM, and routing the prompts to the appropriate model~\citep{Learning2025chuang,Confident2025chuang,SketchofThought2025aytes,RouteLLM2025ong}. For example, RouteLLM~\citep{RouteLLM2025ong} trains a lightweight classifier to classify whether the input prompt is relatively hard or easy, then uses this classifier to route input prompts.
However, deployment of both LRM and LLM can be costly in practice.

In this paper, we introduce a \textbf{superposed deployment} strategy for achieving the ``Goldilocks" balance between slow-thinking LRMs and fast-thinking LLMs.
One key inspiration is 
that \textbf{unlearning overthinking} from LRM
while preserving reasoning is possible. %
Specifically, for linear layers in LRM and LLM, denoted as $W_R$ and $W_B$,
we hypothesize the existence of  $L$ such that $L + W_R$ unlearns overthinking only.
In other words, instead of full difference
$\Delta W = W_B - W_R$,
we can isolate an overthinking-specific 
component $L$, which is hypothesized to be low-rank, by low intrinsic dimensionality~\citep{Measuring2018li}.

With this formulation, our problem reduces 
to finding $L$: Considering both $L$ and $\Delta W$ have similar direction-- moving from $W_R$ to $W_B$-- we derive $L$ by low-rank approximating $\Delta W$ with rank $r$. 
To optimize $r$,
 we introduce an entropy-based, layer-wise rank selection method inspired by~\cite{xuTRPTrainedRank2020}. We show that this layer-wise $r$ selection achieves faster inference and higher task performance compared to na\"ively using uniformly fixed ratio of $r$ across every layer.
The key contributions of \ours are:
\begin{itemize}
    \item We propose \ours, a \textbf{superposed deployment} of slow-thinking LRMs and fast-thinking LLMs with ``Goldilocks" balance between them.
    \item Our key intuition is unlearning just right, or, avoiding overthinking while preserving reasoning, which we enable by low-rank approximating the parameter difference between LRM and LLM.
    \item Experiments with QwQ-32B, DeepSeek-R1-Distill-32B, on ASDIV, GSM8K, AIME, and GPQA show that we draw LRM near the Goldilocks zone-- we can get up to 2.7\texttimes \xspace speedup with negligible performance drop, while reducing the additional GPU memory by 9\texttimes \xspace compared with router-based solutions.
    \item We release our code and models to the public,\footnote{https://anonymous.4open.science/r/GoldSwitch-377D/} enabling further research in this area.
\end{itemize}

%% file: contents/2_rel.tex
\section{Related Works}

\subsection{Overthinking in LRMs}
LRMs like DeepSeek-R1 and QwQ reportedly excel at complex multi-step tasks through chain-of-thought (CoT) training. 
Meanwhile, they may generate long and repeated reasoning paths even for simple problems~\citep{Stop2025sui,NOT2025chen,Danger2025cuadron}. \citet{NOT2025chen} define overthinking as the inefficient use of computation, where models spend too much effort on simple tasks without much accuracy gain. 
For example, they show that even for a simple question like ``What is the answer of 2 plus 3?'', LRM such as QwQ-32B-Preview requires 901 tokens to reach the answer `5', while an LLM from the same family, Qwen2.5-72B, requires only 7 tokens.

\subsection{Reducing Overthinking}
\subsubsection{Training-Based}\label{sec:train}
Reinforcement learning or supervised fine-tuning approaches, by rewarding concise, accurate answers and penalize long or incorrect ones, can encourage the model to generate shorter reasoning paths~\citep{Can2024liu,Distilling2024yu,Stepwise2025cui,SelfTraining2025munkhbat,CoTValve2025ma,C3oT2025kang,DAST2025shen,L12025aggarwal,Optimizing2025qu,Training2025arora,Kimi2025team,O1Pruner2025luo,Demystifying2025yeo,TokenSkip2025xia}. 
For example, models like Kimi k1.5~\citep{Kimi2025team} apply length-aware rewards during their RL process to encourage shorter responses. It also utilizes rejection sampling, generating and evaluating candidate responses to guide efficient generation. Other methods create length-optimized datasets by removing redundant steps or tokens~\citep{TokenSkip2025xia, C3oT2025kang}, or use self-training to sample shorter paths as pseudo-labels~\citep{SelfTraining2025munkhbat}.

Models like Qwen3~\cite{Qwen32025yang} are trained to allow both slow- and fast- thinking by predefined prompts. However, due to their limited performance, the recent trend is back to releasing slow- and fast- thinking models separately~\cite{Qwen32025yang,Qwen3Omni2025xu,qwen3max}.

\paragraph{Distinction}
While these methods usually require extensive data- and compute- resources, \ours is training-free, removing such a burden.

\subsubsection{Training-Free}
\paragraph{Routing-based}
Routing-based method classifies the input prompt to route to an appropriate model~\citep{Learning2025chuang,Confident2025chuang,SketchofThought2025aytes,RouteLLM2025ong}. For example, RouteLLM~\citep{RouteLLM2025ong} trains a lightweight classifier to classify whether the input prompt is relatively hard or easy, then uses this classifier to route input prompts.

\paragraph{Prompt-based}
Prompt-based approaches adjust the input format to limit reasoning steps~\citep{Benefits2024renze,Chain2025xu,Reasoning2025ma}. For example, minimal prompting strategies like Nothinking~\citep{Reasoning2025ma} encourage direct answers without intermediate steps, by early closure of <think> tag with </think> (\autoref{fig:case_nothink}).

\paragraph{Distinction}
While the routing-based method requires deploying both LRM and LLM, we propose a superposed deployment strategy.
Moreover, our method is orthogonal to prompt-based methods, which can be combined to further enhance the efficiency.
\subsection{Low-Rank Representation}
Parameter-Efficient Fine-Tuning (PEFT) methods, especially LoRA~\citep{LoRA2022hu}, have become a standard for adapting large language models (LLMs) to downstream tasks. These methods rely on the hypothesis that the parameter changes required to learn a specific task or capability have a much lower intrinsic dimension than the full parameter space~\citep{LoRA2022hu,Intrinsic2021aghajanyan}. 

MoDeGPT~\cite{MoDeGPT2025lin} modularizes model parameters by task and recombines them, demonstrating that LLMs can be compressed and reused efficiently without additional training. 
\paragraph{Distinction}
\ours demonstrates that unlearning reasoning capacities can also be captured in a compact, low-dimensional subspace without additional training.

%% file: contents/3_prop.tex
\section{Proposed Method}
\subsection{Preliminaries}
\subsubsection{LRMs}
To enhance the reasoning capabilities of large language models (LLMs), researchers have developed various reasoning-oriented models, such as QwQ~\citep{qwq32b} and DeepSeek-R1~\citep{DeepSeekR12025deepseek-ai}. These models are typically trained on a base model (e.g., Qwen2.5-32B) with extensive chain-of-thought (CoT) supervision, enabling them to perform complex reasoning tasks.

Formally, we denote the base model as $W_B$. 
By updating this model we obtain a reasoning-enhanced model $W_R$:
\begin{equation}
W_R = W_B + \Delta W
\end{equation}
where $\Delta W$ represents the parameter update needed for complex reasoning.

However, this reasoning-enhanced model often suffers from overthinking, where it generates unnecessarily long or convoluted reasoning paths~\citep{Stop2025sui}. This behavior can lead to inefficiencies and even degradation of task performance~\citep{Danger2025cuadron}.

\subsubsection{Baseline: Routing Input Queries to LRM and LLMs}\label{sec:prop:routellm}
To avoid overthinking, we can adopt RouteLLM~\citep{RouteLLM2025ong} to route the input query to the appropriate model.

Formally, given a lightweight classifier $C$ that predicts the difficulty of the input, RouteLLM routes the input $x$ to $W_R$ if $C(x) \le t$, and to $W_B$ otherwise.
In theory, the classifier $C$ can be either black-boxed, or a BERT-based classifier trained on a preference dataset.

However, this solution blindly deploys both $W_R$ and $W_B$, which can be costly in practice.

\begin{figure*}[]
\centering
\includegraphics[width=0.8\textwidth]{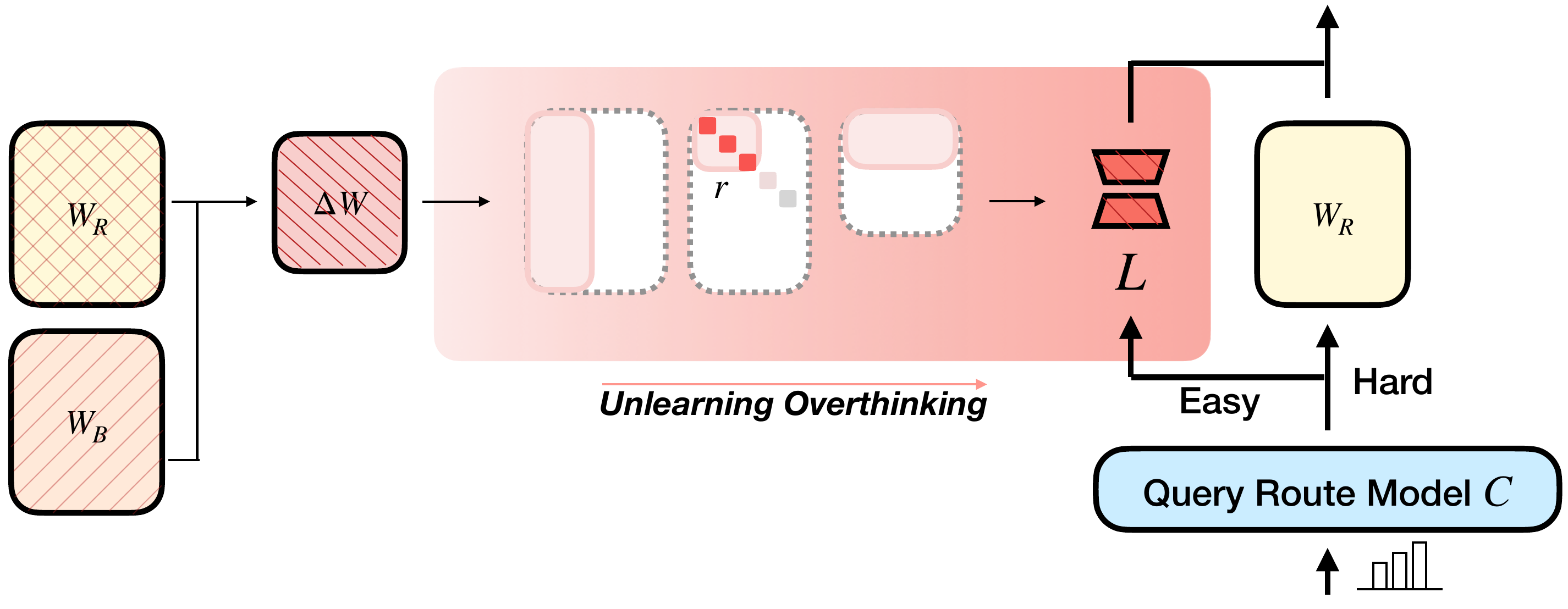}
\caption{
Overview of \ours. \ours extracts the overthinking parameters from the reasoning-enhanced model using low-rank approximation. It then dynamically adjusts the reasoning capabilities based on the difficulty of the input, allowing for efficient reasoning without retraining.
}
\label{fig:overview}
\end{figure*}

\subsection{\ours: Training-Free Superposition Between  LRM and LLM}
\autoref{fig:overview} illustrates the overall flow of \ours. 
We start by defining  $L$, for unlearning only overthinking  from LRM (\S \ref{sec:unlearn}).
This lightweight module could be deployed alongside $W_R$, with minimal overhead. Then we introduce how to superpose $L$ with the given LRM, for each given query (\S \ref{sec:superpose}).

\subsubsection{Training-Free $L$ to Unlearn Only Overthinking from LRM}\label{sec:unlearn}

\paragraph{$L$ as a low-rank approximation of $\Delta{W}$}
The key intuition to construct $L$ is that \textbf{unlearning overthinking} while preserving reasoning is possible. In other words, unlike $\Delta W = W_B - W_R$, which captures all finetuned capabilities including reasoning, $L$ focuses solely on overthinking, implying lower intrinsic dimensionality than $\Delta W$~\citep{Measuring2018li}.
With low intrinsic dimensionality, $L$ could be low-rank decomposed, enabling lightweight deployment alongside $W_R$.

Our second key intuition is that $L$ and $\Delta W$ have similar direction, both shifting $W_R$ to $W_B$.
Therefore we hypothesize that $L$ can be a low-rank approximation of $\Delta W$.
Our experiments (Section \ref{sec:exp}) will reveal that when we low-rank approximate the full unlearning matrix $\Delta W$, slow-thinking unlearning is captured first in these low-rank approximations.

Formally, we decompose $\Delta{W}$ as : 
\begin{equation}
\Delta W \approx U S V^T
\end{equation}

where $U \in \mathbb{R}^{N \times \min(N,M)}$, $S \in \mathbb{R}^{\min(N,M) \times \min(N,M)}$ and $V \in \mathbb{R}^{M \times \min(N,M)}$. Following the Eckart-Young-Mirsky theorem~\citep{approximation1936eckart}, we can approximate $\Delta W$ with a low-rank matrix by selecting the top $r$ singular values from $S$. 
This allows us to select the essential dimensions of the delta matrix $\Delta W$.
Formally, we get the low-rank approximation of $\Delta W$ as follows:

\begin{equation}\label{eq:loW_Rank_approx}
\hat{U} = U[:, :r], \quad \hat{S} = S[:r, :r], \quad \hat{V} = V[:, :r]
\end{equation}

Finally, we get the \ours $L$ as follows:

\begin{equation}
L = L_U \cdot L_V, \quad L_U = \hat{U} \cdot \sqrt{\hat{S}}, \quad L_V = \sqrt{\hat{S}} \cdot \hat{V}^T
\end{equation}

The resulting $L$ can selectively remove unnecessary overthinking components while preserving core reasoning.
This approach is crucial for retaining the reasoning   of large LRMs while enhancing  efficiency during inference.

\paragraph{Energy-based $r$ selection}
 Now, the problem narrows down to how to effectively select $r$ for each layer. A blind selection of a high rank would make $L$ too similar to $\Delta W$, causing the unlearning process to strip away not only overthinking components but also critical reasoning abilities. 

To address this, inspired by~\citet{xuTRPTrainedRank2020}, we employ an energy-based selection mechanism driven by the energy percentage $e$. According to matrix perturbation theory~\citep{stewart1990matrix}, the difference between two matrices is bounded by the differences in their singular values, as expressed by the inequality:

\begin{equation}
\|B-A\|_F^2 \ge \sqrt{\sum_i |s_{b,i} - s_{a,i}|^2}
\end{equation}
where $s_{b,i}, s_{a,i}$ denotes the singular values of $B$ and $A$ respectively. Applying this to our setting, we derive the lower bound of the error between $L$ and $\Delta W$:
\begin{equation}\label{eq:frob}
\|L - \Delta W\|_F ^2 \ge \sqrt{\sum_{i=r}^{\min(N,M)} s_i^2}
\end{equation}
where $s_i$ denotes the singular values of $\Delta W$.
This lower bound serves as an indicator to prevent $L$ from becoming too close to $\Delta W$, ensuring that essential reasoning is not inadvertently discarded. For each layer, we determine the optimal $r$ by regulating the right-hand term of \autoref{eq:frob}, preventing it from diminishing excessively.

To this end, we introduce the reversed cumulative energy $e(r)$, defined as follows:
\begin{equation}
e(r) = \frac{\sum_{i=r}^{\min(N,M)} s_i^2}{\sum_{i=1}^{\min(N,M)} s_i^2}
\end{equation}
We then select $r$ as the minimum rank that maintains $e(r)$ above an empirically tuned threshold $\tau$. This ensures sufficient information retention while eliminating unnecessary overthinking components.

\subsubsection{Hard and Soft Superposition}\label{sec:superpose}
Now the remaining part is how to superpose slow- and fast-thinking LRM with $L$ and $W_R$. We introduce hard or soft superposition.

\input{tables/1_qwq.tex}

\input{tables/2_deepseek.tex}

\paragraph{Hard Superposition}
Given a classifier $C$ that predicts the difficulty of the input (\S \ref{sec:prop:routellm}), 
we can switch between slow-thinking and fast-thinking based on the predicted difficulty:
\begin{equation}
W'_R = \begin{cases}
    W_R + L & \text{if } C(x) \le t \\
    W_R & \text{if } C(x) < t
\end{cases}
\end{equation}
where $W'_R$ is the weight we use for the inference of input $x$, and $t$ is a threshold that determines whether the input is considered easy or hard.
This enables us to deal with the complex tasks that require sophisticated reasoning capabilities via slow-thinking, while still benefiting from the reduced overthinking for easier tasks via fast-thinking.

\paragraph{Soft Superposition}
Moreover, we investigate soft superposition to continuously adjust the weight of \ours based on the difficulty of the input, by generalizing the concept of routing in \citet{RouteLLM2025ong}.

Formally, the softly superposed $W'_r$ is defined as:
\begin{equation}
W'_r = W_R + \alpha L
\end{equation}
where $\alpha$ is a scaling factor that controls the strength of \ours. 
However, we find that scaling \ours may increase the overthinking in some cases, making it harder to control the reasoning capabilities (\autoref{fig:ds_asdiv_avgtok}). Therefore we use the hard superposition for the final experiments.

%% file: tables/1_qwq.tex
\begin{table*}[]
\centering
\begin{tabular}{l|c|cc|cc}
\hline
                                           & Relative Size  & GSM8K  & time           & ASDIV  & time           \\ \hline
QwQ-32B                                    & 1.00\texttimes & 96.2\% & 1.00\texttimes & 87.6\% & 1.00\texttimes \\
Qwen2.5-Instruct-32B                       & 1.00\texttimes & 94.2\% & 0.17\texttimes & 87.6\% & 0.22\texttimes \\
RouteLLM                                   & 2.00\texttimes & 93.8\% & 0.61\texttimes & 88.3\% & 0.37\texttimes \\
QwQ-32B + \ours always on                    & 1.11\texttimes & 95.5\% & 0.18\texttimes & 87.9\% & 0.25\texttimes \\
QwQ-32B + \ours                              & 1.11\texttimes & 95.8\% & 0.37\texttimes & 87.5\% & 0.39\texttimes \\
QwQ-32B + NoThinking~\citep{Reasoning2025ma} & 1.00\texttimes & 95.9\% & 0.98\texttimes & 87.2\% & 1.06\texttimes \\ \hline
\end{tabular}
\caption{Comparison on GSM8K and ASDIV with QwQ-32B.}
\label{tab:qwq}
\end{table*}

%% file: tables/2_deepseek.tex
\begin{table*}[]
\centering
\begin{tabular}{l|c|cc|cc}
\hline
                                                    & Relative Size  & GSM8K & time           & ASDIV & time           \\ \hline
DS-R1-32B                        & 1.00\texttimes & 88.7\% & 1.00\texttimes & 83.7\% & 1.00\texttimes \\
Qwen2.5-32B                                         & 1.00\texttimes & 85.5\% & 0.91\texttimes & 80.7\% & 1.03\texttimes \\
RoutLLM                                             & 2.00\texttimes & 88.6\% & 0.84\texttimes & 83.0\% & 0.81\texttimes \\
DS-R1-32B + \ours                & 1.11\texttimes & 92.3\% & 0.85\texttimes & 86.0\% & 0.82\texttimes \\
DS-R1-32B + NoThinking           & 1.00\texttimes & 89.2\% & 0.54\texttimes & 87.1\% & 0.44\texttimes \\
DS-R1-32B + \ours +   NoThinking & 1.11\texttimes & 89.9\% & 0.57\texttimes & 86.6\% & 0.48\texttimes \\ \hline
\end{tabular}
\caption{Comparison on GSM8K and ASDIV with DeekSeek-R1-Distill-Qwen-32B.}
\label{tab:deepseek}
\end{table*}

%% file: contents/4_exp.tex
\section{Experiments}\label{sec:exp}

In this section, we aim to address the following research questions:
\begin{itemize}
\item RQ1: Does \ours effectively reduce overthinking in LRMs?
\item RQ2: Does \ours effectively retain the complex reasoning capability of LRMs?
\item RQ3: Is \ours cost-effective to build and serve?
\item RQ4: Is the entropy-based rank selection method effective? 
\item RQ5: How does soft switching compare to hard switching?
\end{itemize}
\input{tables/2_qwen3.tex}
\begin{figure*}[]
\includegraphics[width=\textwidth]{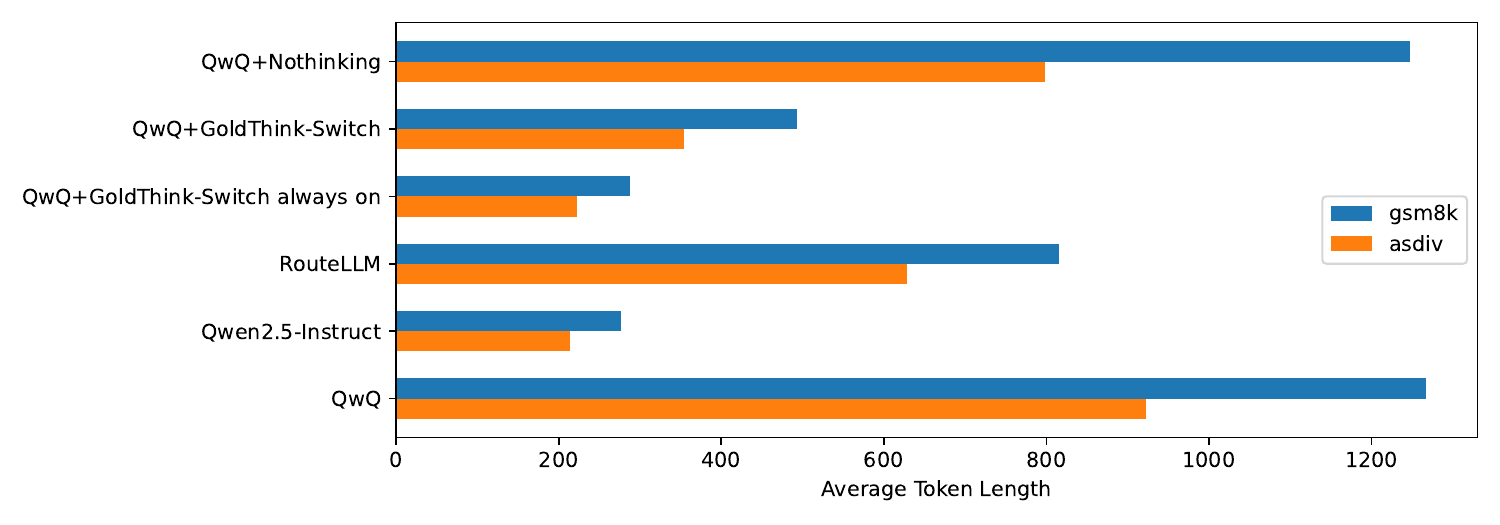}
\caption{Average token length of GSM8K and ASDIV on QwQ}\label{fig:qwq_avgtok}
\end{figure*}

\paragraph{Models}
We employ three open-sourced LRMs: QwQ-32B~\citep{qwq32b}, DeekSeek-R1-Distill-Qwen-32B~\citep{DeepSeekR12025deepseek-ai}, and Qwen3-4B-Thinking-2507~\cite{Qwen32025yang}.
As the base LLMs, we utilize Qwen2.5-32B-Instruct, Qwen2.5-32B~\citep{Qwen22025qwen}, and Qwen3-4B-Instruct-2507, respectively.

\paragraph{Tasks and Datasets}
We mainly evaluate with ASDIV~\citep{ASDIVDiverse2020miao} and GSM8K~\citep{GSM8KTraining2021cobbe}, where overthinking is reported to happen~\citep{NOT2025chen}. As challenging reasoning problems where slow thinking is favored, we employ AIME2024. We also include GPQA~\citep{GPQA2024rein}, the non-math task \citet{NOT2025chen} employed. All evaluations are done in a 0-shot manner.

\paragraph{Implementation Details}
For AIME2024 and GPQA, we employ \textsc{open-r1}.\footnote{github.com/huggingface/open-r1} To evaluate ASDIV and GSM8K, we extend \textsc{lm-evaluation-harness}\footnote{github.com/neuralmagic/lm-evaluation-harness/tree/llama\_3.1\_instruct}~\citep{eval-harness} to use the suggested format by DeepSeek and Qwen. Following the suggestion, we use temperature of 0.6, topP of 0.95, topK of 30, and minP of 0. 
For Qwen3 models, we use topK of 20, and for Qwen3-4B-Instruct-2507 and the non-thinking mode of Qwen3-4B, we use temperature of 0.7, topP of 0.8.
We use max new tokens of 32K, while 
for ASDIV and GSM8K evaluation of DeepSeek-R1-Distill-Qwen-32B and Qwen2.5-32B, we use max new tokens of 2K, to align with the default configuration of huggingface Qwen2.5-32B. We use $t=0.78$, and $\tau=0.6$.\footnote{We conducted small grid search once on QwQ-32B with 100 samples of GSM8K to choose this value, and used over all models and tasks, as elaborated in Appendix B.} 
For classificer $C$, we use a pretrained BERT-based model introduced in \citet{RouteLLM2025ong}, which is claimed to generalize over diverse settings.
All experiments are conducted on 2\texttimes A6000.

\paragraph{Comparisons}
We compare the following methods:
\begin{itemize}
    \item \textbf{LRM} breaks down structured problems into multiple steps for deeper understanding but often overthink even on simple tasks. 
    \item \textbf{LLM} can produce concise responses with low computation, but often lacks deep logical reasoning for complex problems.
    \item \textbf{RouteLLM}~\citep{RouteLLM2025ong} selects different language models (such as LLM or LRM) depending on the difficulty of the input. For simple inputs, it chooses a lightweight LLM, and for harder problems, it uses an LRM. In our setup, we use a BERT-based router with a confidence threshold to decide which model to use, which we use in \ours as well.
    \item \textbf{Nothinking}~\citep{Reasoning2025ma} skips the reasoning process by pre-filling an assistant message with '</think>' without training. It also uses a Token Budget (set to 3500 in our tests) to force short responses, reducing overthinking. 
    \item \textbf{\ours} is our proposal. We define the three methods as follows. 
    \begin{itemize}
        \item \textbf{Always On} applies unlearning reasoning to all queries without distinguishing task difficulty.
        \item \textbf{Hard Switch} completely enables or disables unlearning based on a query classifier $C$.
    \item \textbf{Soft Switch} adjusts unlearning based on query difficulty, providing continuous control.
    \end{itemize}
\end{itemize}

\input{tables/3_aime.tex}

\input{tables/3_aime_deepseek.tex}

\subsection{RQ1: \ours Effectively Reduces Overthinking in LRMs}\label{sec:exp:rq1}
\autoref{tab:qwq}, \ref{tab:deepseek} and \ref{tab:qwen3} show the results of \ours on QwQ-32B, DeepSeek-R1-Distill-Qwen-32B, and Qwen3-4B, respectively.
Taking QwQ-32B as an example, on GSM8K, while QwQ-32B achieves higher accuracy compared with Qwen2.5-32B-Instruct, it takes almost 6\texttimes \xspace longer time to finish the same task.
By always turning on \ours, we can achieve almost the same speed while reducing the performance gap by 65\%.
Meanwhile, switching on/off \ours further reduces the performance gap, though incurring a moderate time cost.
On ASDIV, we can observe a similar trend-- the performance gap is almost similar, while the inference time is drastically reduced.

\autoref{fig:qwq_avgtok} shows the average token length of GSM8K and ASDIV on QwQ-32B. We can observe that always turning on \ours can significantly reduce the average token length, almost being similar to the base LLM Qwen2.5-32B-Instruct. More detailed statistics are provided in the Appendix.

When compared to an existing router-based solution RouteLLM, \ours achieves similar or better performance, while drastically decreasing the GPU memory usage. For example, on GSM8K, \ours improves the accuracy by 1.7-3.7\% (\autoref{tab:qwq},\ref{tab:deepseek},\ref{tab:qwen3}), while requiring similar or only 0.6\texttimes \xspace of time to finish (\autoref{tab:qwq}).

When compared to a prompt-based baseline NoThinking, while NoThinking often fails to decrease the generation time (\autoref{tab:qwq} and \ref{tab:qwen3}), \ours always succeeds to decrease the time drastically, up to 2.7\texttimes \xspace (\autoref{tab:qwq}).
Moreover, even when NoThinking is working exceptionally well, \ours can be combined with Nothinking to explore better performance (\autoref{tab:deepseek}).

Finally, when compared to Qwen3-4B, which is trained to have slow- or fast- thinking switch, training-free superposition of Qwen3-4B-Thinking and Qwen3-4B-Instruct by \ours is much better-- GSM8K accuracy is 1.9\% higher, while ASDIV speed is 30\% faster (\autoref{tab:qwen3}).

\input{tables/3_aime_qwen3.tex}

\input{tables/8_build_cost.tex}
\input{tables/8_ttft.tex}

\input{tables/5_entropy.tex}
\subsection{RQ2: \ours Effectively Retains the Complex Reasoning Capability of LRMs}
\autoref{tab:aime_qwq}, \ref{tab:aime_deepseek} and \ref{tab:aime_qwen3} present the results of \ours on AIME2024 and GPQA with QwQ-32B, DeepSeek-R1-Distill-Qwen-32B, and Qwen3-4B, respectively. While always switching on \ours effectively reduces overthinking and preserves performance in easy tasks, it fails to do so in complex tasks. For instance, QwQ-32B achieves 66.7\% accuracy on AIME2024, whereas always enabling \ours drops it to 16.7\%. In contrast, \ours with its switching mechanism retains QwQ-32B's reasoning capability even in complex scenarios, highlighting its necessity. 
The trend on GPQA is consistent as well.

\subsection{RQ3: \ours Is Cost-Efficient to Build and Serve}
\autoref{tab:build_cost} shows that the building cost of \ours is negligible-- this training-free preprocessing step requires less than 1 hour with only 4GB of GPU memory.
\autoref{tab:ttft} shows that the routing time is negligible as well. Even if we consider the time-to-first-token latency, routing time takes only 0.04\texttimes.

Finally, the `Relative Size' column from \autoref{tab:qwq}, \ref{tab:deepseek} and \ref{tab:qwen3} shows that \ours significantly saves the deployment cost. While the routing-based solution, RouteLLM additionally requires 1\texttimes \xspace compared with serving one LRM, \ours only requires around 0.11\texttimes, which is 9\texttimes \xspace smaller.

\subsection{RQ4: Entropy-Based Rank Selection Is Indeed Better}

\autoref{tab:L_size_QwQ} shows the performance of \ours with different rank selection methods on QwQ-32B.
We can observe that entropy-based rank selection can obtain not only a smaller size compared to the fixed ratio method, but also achieves better performance.
This emphasizes the importance of our layer-wise energy-based method to keep distance from $\Delta W$ (Equation \ref{eq:frob}).

\subsection{RQ5: Soft Switching Fails to Reduce Overthinking}

\begin{figure}[]
\centering
\includegraphics[width=\columnwidth]{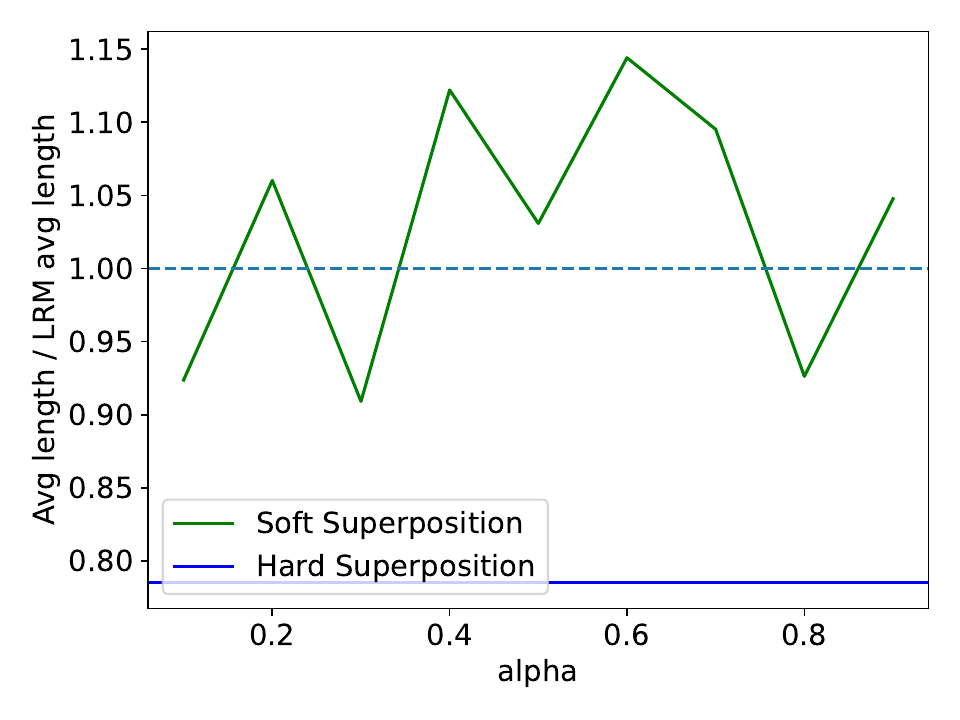}
\caption{Average token length ratio of ASDIV on DeepSeek-R1-Distill-Qwen-32B}\label{fig:ds_asdiv_avgtok}
\end{figure}

\autoref{fig:ds_asdiv_avgtok} illustrates the relative average token length ratio of hard and soft superposition compared to LRM (DeepSeek-R1-Distill-Qwen-32B), using 90 samples from ASDIV. As $\alpha$ varies between 0 and 1, the average token length of output from $W_R + \alpha L$ may increase, suggesting that the soft switching mechanism is ineffective in mitigating overthinking. In contrast, for easy problems, hard superposition effectively reduces token length.

\subsection{Case Study}
\autoref{fig:case_qwq}, \ref{fig:case_nothink} and \ref{fig:case_optthink} in the Appendix show a case study for the same question
``Sophia and Rose went together to the market to buy onions and potatoes. Rose bought 4 times the number of onions and potatoes Sophia bought. If Rose bought 12 onions and 4 potatoes, how many onions and potatoes in total did Sophia buy at the market?''.

First, the base LRM, QwQ-32B, overthinks over and over again-- repeatedly obtaining the same answer ``4'' for 10 times-- resulting in 1353 tokens to reach the final answer.
NoThinking~\citep{Reasoning2025ma}, by closing the </think> beforehand, reduces the thinking process, but fails to avoid repeated answer of ``4'' for 6 times, resulting in 984 tokens to reach the final answer.

In contrast, \ours directly dives into a single CoT process, without any repetitive answering, resulting in 328 tokens, almost 4\texttimes \xspace reduction of token length compared to the base LRM. Moreover, it is interesting to observe that %
the LRM seems to forget the <think> token, not reacting at all, even does not close with </think> token. 
The output token length is almost similar to the fast-thinking LLM, Qwen2.5-32B-Instruct (\autoref{fig:case_qwen}).

%% file: tables/2_qwen3.tex
\begin{table*}[]
\centering
\begin{tabular}{l|c|cc|cc}
\hline
                                    & Relative Size  & GSM8K & time           & ASDIV & time           \\ \hline
Qwen3-Thinking-4B                   & 1.00\texttimes & 95.2\% & 1.00\texttimes & 89.5\% & 1.00\texttimes \\
Qwen3-Instruct-4B                   & 1.00\texttimes & 90.4\% & 0.30\texttimes & 77.6\% & 0.40\texttimes \\
RoutLLM                             & 2.00\texttimes & 91.4\% & 0.47\texttimes & 78.2\% & 0.53\texttimes \\
Qwen3-Thinking-4B + \ours always on & 1.17\texttimes & 93.9\% & 0.48\texttimes & 88.5\% & 0.56\texttimes \\
Qwen3-Thinking-4B + \ours           & 1.17\texttimes & 93.9\% & 0.47\texttimes & 88.5\% & 0.56\texttimes \\
Qwen3-Thinking-4B + NoThinking      & 1.00\texttimes & 94.7\% & 1.06\texttimes & 89.6\% & 1.07\texttimes \\
Qwen3-4B                            & 1.00\texttimes & 92.0\% & 0.48\texttimes & 88.4\% & 0.81\texttimes \\ \hline
\end{tabular}
\caption{Comparison on GSM8K and ASDIV with Qwen3-4B series.}
\label{tab:qwen3}
\end{table*}

%% file: tables/3_aime.tex
\begin{table}[]
\centering
\setlength{\tabcolsep}{1pt}
\begin{tabular}{l|rl}
\hline
                          & \multicolumn{1}{l}{AIME} & GPQA   \\ \hline
QwQ                       & 66.7\%                   & 63.1\% \\
Qwen2.5-Instruct          & 13.3\%                   & 45.5\% \\
RouteLLM                  & 66.7\%                   & 62.1\% \\
QwQ+Gold-Switch always on & 16.7\%                   & 42.9\% \\
QwQ+Gold-Switch           & 66.7\%                   & 61.1\% \\ \hline
\end{tabular}
\caption{AIME, GPQA performance of \ours and baselines on QwQ-32B.}
\label{tab:aime_qwq}
\end{table}

%% file: tables/3_aime_deepseek.tex
\begin{table}[]
\centering
\setlength{\tabcolsep}{1pt}
\begin{tabular}{l|rl}
\hline
                                & \multicolumn{1}{l}{AIME} & GPQA   \\ \hline
DS-R1-32B                       & 60.0\%                    & 65.1\% \\
Qwen2.5                         & 6.7\%                     & 39.4\% \\
RouteLLM                        & 60.0\%                    & 55.6\% \\
DS-R1-32B+Gold-Switch always on & 23.3\%                    & 43.4\% \\
DS-R1-32B+Gold-Switch           & 60.0\%                    & 56.6\% \\ \hline
\end{tabular}
\caption{AIME, GPQA performance of \ours and baselines on DeepSeek-R1-Distill-Qwen-32B.}
\label{tab:aime_deepseek}
\end{table}

%% file: tables/3_aime_qwen3.tex
\begin{table}[]
\centering
\setlength{\tabcolsep}{1pt}
\begin{tabular}{l|rl}
\hline
                                    & AIME  & GPQA  \\ \hline
Q3T-4B                   & 76.7\% & 66.7\% \\
Q3I-4B                   & 66.7\% & 44.9\% \\
RouteLLM                             & 76.7\% & 58.6\% \\
Q3T-4B + \ours always on & 70.0\% & 57.1\% \\
Q3T-4B + \ours           & 76.7\% & 63.6\% \\
Q3-4B                            & 76.7\% & 53.0\% \\ \hline
\end{tabular}
\caption{AIME, GPQA performance of \ours and baselines on Qwen3-4B (Q3-4B), Qwen3-Thinking-4B (Q3T-4B), and Qwen3-Instruct-4B (Q3I-4B).}
\label{tab:aime_qwen3}
\end{table}

%% file: tables/8_build_cost.tex
\begin{table}[]
\centering
\begin{tabular}{l|cc}
\hline
Model size    & GPU mem (GB) & GPU hours (h) \\ \hline
Q3T-4B  &  1            & 0.06              \\
QwQ-32B & 4            & 0.67          \\ \hline
\end{tabular}
\caption{Required GPU memory (GB) and hours (h) to build \ours with one A6000.}
\label{tab:build_cost}
\end{table}

%% file: tables/8_ttft.tex
\begin{table}[]
\centering
\begin{tabular}{l|cc}
\hline
\textbf{} & Total TTFT (s) & Routing Time (s) \\ \hline
GSM8K     & 0.2348         & 0.0098           \\
ASDIV     & 0.2125         & 0.0070           \\ \hline
\end{tabular}
\caption{Time comparison between the total time-to-first-token (TTFT) and the routing time on QwQ-32B with \ours.}
\label{tab:ttft}
\end{table}

%% file: tables/5_entropy.tex
\begin{table*}[]
\centering
\begin{tabular}{l|c|cc|cc}
\hline
Rank $r$ selection  & relative size of $L$ & GSM8K & time(h) & ASDIV & time (h)\\ \hline
fixed ratio & 14\%                         & 90.8\% & 6.2 &   85.7\%    &  7.5  \\
entropy-based & 11\%                       & 92.3\% & 3.9 &   86.0\%    &  6.3   \\ \hline
\end{tabular}
\caption{The size of $L$ and the performance of GSM8K and ASDIV on DeepSeek-R1-Distill-Qwen-32B}
\label{tab:L_size_QwQ}
\end{table*}

%% file: contents/5_concl.tex
\section{Conclusion}
Our work 
introduces lightweight, training-free LoRA modules that suppress overthinking, enabling LLMs to dynamically modulate their inference budget based on input complexity. 
This approach not only avoids the need for deploying two models, but also makes inference more efficient.

\section{Limitations}

\ours leverages classifiers to route hard or easy tasks from existing baselines~\cite{RouteLLM2025ong}. Therefore we follow the limitations from those baselines, such as interpretability, or scarcity of gold-labeled dataset to train such classifiers.

\ours assumes access to both the base and fine-tuned weights, which may not be feasible for closed-source or commercial LLMs. However, our target is making local deployment more efficient with superposed deployment, thus supporting the closed-source models is out of our scope.

%% file: contents/7_appendix.tex
\section{Token-Length Statistics}
\input{tables/9_gsm8k_tokens.tex}
\input{tables/9_asdiv_tokens.tex}
\autoref{tab:token_length_gsm8k}, \ref{tab:token_length_asdiv} show the detailed statistics of token-length distribution.

\section{Ablation Study} %
\paragraph{Effect of Thresholds}
We conducted ablation study to investigate how the performance of \ours varies with different values of the difficulty threshold $t$ and energy threshold $\tau$.
\input{tables/7_threshold}
As shown in Table~\ref{tab:threshold}, we empirically selected the best hyperparameters based on performance across a small grid search.

\input{tables/case_qwq.tex}
\input{tables/case_nothinking.tex}
\input{tables/case_optthinkswitch.tex}
\input{tables/case_qwen.tex}

%% file: tables/9_gsm8k_tokens.tex
\begin{table*}[]
\centering
\setlength{\tabcolsep}{3pt}
\begin{tabular}{l|ccccccc}
\hline
\textbf{}                      & mean    & min & max   & var      & Q(0.25) & Q(0.5) & Q(0.75) \\ \hline
QwQ+Nothinking                 & 1342.59 & 58  & 14261 & 1.43E+06 & 750            & 1025          & 1487.75        \\
QwQ+GoldThink-Switch always on & 285.61  & 108 & 749   & 7.17E+03 & 224            & 275           & 335            \\
QwQ+GoldThink-Switch           & 494.00  & 110 & 9966  & 3.99E+05 & 233            & 298           & 419.75         \\
Qwen2.5-Instruct               & 275.82  & 110 & 796   & 5.67E+03 & 222            & 266           & 318            \\
QwQ-32B                        & 1268.87 & 285 & 9484  & 1.11E+06 & 697.5          & 968           & 1397.75        \\ \hline
\end{tabular}
\caption{Token-length statistics on GSM8K with QwQ-32B. We report mean, min, max, variance (var), quantiles (Q).}
\label{tab:token_length_gsm8k}
\end{table*}

%% file: tables/9_asdiv_tokens.tex
\begin{table*}[]
\centering
\setlength{\tabcolsep}{3pt}
\begin{tabular}{l|ccccccc}
\hline
\textbf{}                      & mean   & min & max   & var      & Q(0.25) & Q(0.5) & Q(0.75) \\ \hline
QwQ+Nothinking                 & 918.94 & 41  & 32768 & 3.34E+06 & 229            & 660           & 1025           \\
QwQ+GoldThink-Switch always on & 221.80 & 79  & 32768 & 4.70E+05 & 141            & 182           & 245            \\
QwQ+GoldThink-Switch           & 351.98 & 77  & 32768 & 7.95E+05 & 144            & 189           & 291            \\
Qwen2.5-Instruct               & 212.77 & 79  & 1113  & 8.36E+03 & 157            & 187           & 241            \\
QwQ-32B                        & 921.37 & 225 & 12999 & 7.81E+05 & 501            & 680           & 1005           \\ \hline
\end{tabular}
\caption{Token-length statistics on ASDIV with QwQ-32B. We report mean, min, max, variance (var), quantiles (Q).}
\label{tab:token_length_asdiv}
\end{table*}

%% file: tables/7_threshold.tex
\begin{table}[H]
\centering
\begin{tabular}{ccccc}
\hline
$\tau$ $\backslash$ $t$    & 0.76 & 0.77 & 0.78 & 0.79 \\ \hline
0.6 & 93\% & 92\% & \textbf{95\%} & 91\% \\
0.7 & 93\% & 93\% & 93\% & 92\% \\ \hline
\end{tabular}
\caption{Comparison on GSM8K with Qwen3-4B }
\label{tab:threshold}
\end{table}

%% file: tables/case_qwq.tex
\begin{figure*}
\fbox{
\begin{minipage}{\textwidth}
{\color{orange}<think>}
Okay, let's see. So the problem is about Sophia and Rose buying onions and potatoes at the market. Rose bought 4 times the number of onions and potatoes that Sophia bought. And we know Rose bought 12 onions and 4 potatoes. The question is asking how many onions and potatoes Sophia bought in total.

Hmm, first, \textbf{let me parse the problem again.} (…omitted…)
Therefore, Sophia bought 3 onions and 1 potato. \textit{The total would be 3 + 1 = 4}. 

\textbf{Wait, but let me make sure I didn't misinterpret the problem} (…omitted…)

\textbf{Alternatively,} (…omitted…) \textbf{Alternatively,} (…omitted…)
Therefore, \textit{total for Sophia is 3 + 1 = 4}. 

\textbf{Alternatively,} (…omitted…) \textit{Sophia's total is 4}. 

Hmm, so either way, whether we take it per item or combined, the answer comes out the same? Wait, that's interesting. \textbf{Let me check both interpretations again.} (…omitted…)

So either way, \textit{the answer is 4}. \textbf{Wait,} (...omitted...) Therefore, \textit{the answer is 4}.

Therefore, I think \textit{the answer is 4}. \textbf{Let me just confirm once again.} (…omitted…)
So Sophia's \textit{total is 4}. 

\textbf{Alternatively,} (…omitted…)
Therefore, Sophia bought 3 onions and 1 potato, \textit{totaling 4. }

I think that's solid. So \textit{the answer should be 4.}
{\color{orange}</think>}

Sophia bought onions and potatoes such that Rose, who purchased 4 times as many of each, ended up with 12 onions and 4 potatoes. Let \( O \) be the number of onions and \( P \) the number of potatoes Sophia bought. Since Rose bought 4 times as many:

\[
4O = 12 \implies O = \frac{12}{4} = 3
\]
\[
4P = 4 \implies P = \frac{4}{4} = 1
\]

Thus, Sophia bought \textit{\( 3 + 1 = 4 \) items in total.}
\boxed{4} (1353 tokens)
\end{minipage}
}
\caption{Case study of QwQ}\label{fig:case_qwq}
\end{figure*}

%% file: tables/case_nothinking.tex
\begin{figure*}
\fbox{
\begin{minipage}{\textwidth}
{\color{orange}<think>}
Okay I have finished thinking.{\color{orange}</think>}

Alright, let's tackle this problem step by step. So Sophia and Rose went to the market to buy onions and potatoes. The problem says that Rose bought 4 times the number of onions and potatoes that Sophia bought. Then it tells us that Rose bought 12 onions and 4 potatoes. We need to find out how many onions and potatoes Sophia bought in total. Hmm, let me break this down. (...omitted...)

Adding those together, the \textit{total would be 3 + 1 = 4. }

\textbf{Wait, let me check that again to make sure I didn't make a mistake.} (…omitted…)
Adding those together, \textit{3 + 1 is indeed 4}. That seems straightforward. 

\textbf{Is there another way to approach this? Maybe by setting up equations.} (…omitted…)
\textit{Total for Sophia is O\_s + P\_s = 3 +1 =4}. Yep, same answer. 

\textbf{Hmm, maybe I should check if there's any possibility that} (…omitted…) the \textit{final total for Sophia ends up being 4}. 

\textbf{But the problem says} "4 times the number of onions and potatoes", which is a bit ambiguous. (…omitted…)

\textbf{Alternatively,} (…omitted…)
So either way, the answer comes out the same. 

Therefore, regardless of the interpretation, the total number of onions and potatoes \textit{Sophia bought is 4}. 

So the \textit{final answer should be 4}. I think that's solid. Both interpretations lead to the same result, so there's no ambiguity here. 

**Final Answer**
The total number of onions and potatoes \textit{Sophia bought is \boxed{4}}. (984 tokens)
\end{minipage}
}
\caption{Case study of QwQ+Nothinking}
\label{fig:case_nothink}
\end{figure*}

%% file: tables/case_optthinkswitch.tex
\begin{figure*}
\fbox{
\begin{minipage}{\textwidth}
{\color{orange}<think>}
Let's break down the problem step by step. (...omitted...)
Therefore, the total number of onions and potatoes that Sophia bought is \(\boxed{4}\). (328 tokens)
\end{minipage}
}
\caption{Case study of QwQ+\ours}\label{fig:case_optthink}
\end{figure*}

%% file: tables/case_qwen.tex
\begin{figure*}
\fbox{
\begin{minipage}{\textwidth}
To find out how many onions and potatoes Sophia bought, (...omitted...) Therefore, Sophia bought a total of 4 items (onions and potatoes) at the market. (273 tokens)
\end{minipage}
}
\caption{Case study of Qwen2.5-32B-Instruct}\label{fig:case_qwen}
\end{figure*}